\documentclass[journal]{IEEEtran}
\usepackage{amsmath}
\usepackage{amsfonts}
\usepackage{diagbox}
\usepackage{multirow}
\usepackage{algorithm}
\usepackage{algorithmicx}
\usepackage{algpseudocode}
\usepackage{graphicx}
\usepackage{subfigure}
\usepackage{makecell}
\usepackage{url}

\ifCLASSINFOpdf
\else
\fi
\begin{document}
%
\title{ScoreGrad: Multivariate Probabilistic Time Series Forecasting with Continuous Energy-based Generative Models}
%
%
%

\author{Tijin Yan,
        Hongwei Zhang,
        Tong Zhou,
        Yufeng Zhan
        and~Yuanqing Xia,~\IEEEmembership{Senior Member,~IEEE}
\thanks{Y. Xia, T. Yan, H. Zhang, T. Zhou and Y. Zhan are with the School of Automation, Beijing
	Institute of Technology, Beijing 100081, China.}
\thanks{Corresponding author: Yuanqing Xia, Email: xia\_yuanqing@bit.edu.cn.}
}

\maketitle

\begin{abstract}
Multivariate time series prediction has attracted a lot of attention because of its wide applications such as intelligence transportation, AIOps. Generative models have achieved impressive results in time series modeling because they can model data distribution and take noise into consideration. However, many existing works can not be widely used because of the constraints of functional form of generative models or the sensitivity to hyperparameters. In this paper, we propose ScoreGrad, a multivariate probabilistic time series forecasting framework based on continuous energy-based generative models. ScoreGrad is composed of time series feature extraction module and conditional stochastic differential equation based score matching module. The prediction can be achieved by  iteratively solving reverse-time SDE. To the best of our knowledge, ScoreGrad is the first continuous energy based generative model used for time series forecasting. Furthermore, ScoreGrad achieves state-of-the-art results on six real-world datasets. The impact of hyperparameters and sampler types on the performance are also explored. Code is available at \url{https://github.com/yantijin/ScoreGradPred}. 
\end{abstract}


%
\IEEEpeerreviewmaketitle

\section{Introduction}
%
%
%
%
\IEEEPARstart{S}{ystems} in various domains become more and more complex in modern societies.  Fortunately, a wide range of sensors can be applied to record the status of systems. As there may be correlations between the states of a system, the recorded data from different sensors can form multivariate time series data. In order to model the evolution of states of complex systems, great efforts have previously been made for multivariate time series prediction. It has attracted much attention because of its wide applications in various fields such as intelligence transportation systems \cite{lana2018road},  IT operations \cite{alexandrov2019gluonts}. With the development of deep learning in recent years, much progress has been made for multivariate time series forecasting.

Despite the advances in time series forecasting tasks, existing methods have some limitations. For example, some methods \cite{su2019robust,shih2019temporal} can not model stochastic information in time series. Furthermore,  some works \cite{salinas2020deepar,wang2019deep} can not model long range time dependencies. Recently, generative models are proved effective for sequential modeling. \cite{oord2016wavenet} proposes TCN with dilation to model long range dependencies and constructs a generative model called WaveNet. \cite{rasul2020multi} combines transformer and masked autoregressive flow together and achieve impressive results. However, the functional form of the models based on VAE and flow based models are constrained. TimeGrad \cite{rasul2021autoregressive} uses an energy-based generative model (EBM) which transforms data distribution to target distribution by slowly injecting noise and achieves state of the art results on many datasets.

Although the energy-based generative model used in TimeGrad is less restrictive on functional forms compared with VAE and flow based models, it still has some limitations. Firstly, denoising diffusion probabilistic models (DDPM) used in TimeGrad is sensitive to the noise scales injected to the original distribution. Secondly, the number of steps used for noise injection needs to be carefully designed. Thirdly, the sampling methods for generation process in DDPM can be further extended.

In order to address these issues,  we propose ScoreGrad, a general framework for multivariate time series forecasting based on continuous energy-based generative models. DDPM can be regarded as discrete form of a stochastic differential equation (SDE) as noted by \cite{song2020score}. Based on this assumption, the number of steps can be replaced by the interval of integration. Furthermore, the noise scales can be easily tuned by diffusion term in SDE. In the end, a conditional continuous energy-based generative model combined with sequential modeling methods is established for forecasting tasks. The prediction can be achieved by iteratively sampling from the reverse continuous-time SDE. Any numerical solvers for SDEs can be used for sampling.

All in all, the contributions of this paper can be summarized as

\begin{enumerate}
\item To the best of our knowledge, ScoreGrad is the first continuous energy-based generative model used for multivariate time series forecasting. 

\item A general framework based on continuous energy-based generative models for time series forecasting is established. The training process at each step is composed of a time series feature extraction module and a conditional SDE based score matching module. The prediction can be achieved by solving reverse time SDE.

\item ScoreGrad achieves state-of-the-art results for multivariate time series forecasting on six real-world datasets with thousands of correlated dimensions.

\end{enumerate}

\section{Related work}
In this section, we firstly review some methods for sequential modeling and multivariate time series forecasting. Then energy-based generative models, especially score matching models are briefly introduced.
\subsection{Multivariate time series forecasting}
Time series forecasting has been studied for a long time and the approaches can be split into a few categories. The statistical modeling methods such as exponential average, moving average, ARIMA \cite{box1968some} and Gaussian process regression have been well established and applied to industrial production. More recently, deep learning based methods have achieved promising results by using the output of statistical engines as features. DeepAR \cite{salinas2020deepar} combines traditional AR models with RNNs by modeling a probabilistic distribution in an autoencoder fashion. MQRNN \cite{wen2017multi} uses MLPs as a decoder to solve the error accumulation issue in DeepAR.
 Furthermore, a few works explore the combination of RNNs with attention, residual connection and dilation \cite{chang2017dilated,kim2017residual,qin2017dual}. Recently, probabilistic models which explicitly model the data distribution based on generative models like normalizing flows \cite{de2020normalizing}, GANs \cite{yoon2019time} achieve better performance than those of deterministic modeling methods. However, the function forms of these methods are constrained and some methods are sensitive to hyperparameters. In this paper, an auto-regressive forecasting framework based on probabilistic modeling framework is established in ScoreGrad.

\subsection{Energy-based generative models}
Energy-based models are un-normalized probabilistic models which specify probability density or mass function up to an unknown normalizing constant \cite{song2021train}. Compared with VAE \cite{kingma2013auto} and some flow-based generative models \cite{dinh2014nice,rezende2015variational}, EBMs directly estimate the unnormalized negative log-probability and do not place a restriction on the tractability of the normalizing constants. Therefore, EBMs are much less restrictive in functional form and have wide applications in various fields such as natural language processing \cite{mikolov2013distributed, deng2020residual}, density estimation \cite{wenliang2019learning,song2020sliced}.

Although EBMs can provide significant modeling advantages, the unknown normalizing constant of EBMs will make training particularly difficult. There are currently several methods for training EBMs: 
\begin{enumerate}
	\item Maximum likelihood estimation based on MCMC. \cite{younes1999convergence} proposes to estimate the gradient of the log-likelihood with MCMC sampling methods such as Hamiltonian Monte Carlo \cite{duane1987hybrid} instead of directly computing the likelihood.
	\item Score matching based methods.  \cite{hyvarinen2005estimation} proposes to minimize a discrepancy between the gradient of the log-likelihood of data distribution and estimated distribution with Fisher divergence. However, the optimization is computationally expensive due to the difficulty of computing the Hessian of log-density functions.
	\item Noise contrastive estimation. \cite{gutmann2010noise} is based on the idea that an EBM can be learned by contrasting it with known density. In practice, proper noise distribution is critical to the success of noise contrastive estimation.
\end{enumerate}

In this paper, we focus on score matching based EBMs, which have achieved state-of-the-art results \cite{ho2020denoising,song2020score} in image generation tasks without using adversarial training methods. The closest related work to ScoreGrad is \cite{rasul2021autoregressive}, which uses denoising diffusion probabilistic models in probabilistic time series forecasting. However, it's sensitive to the noise scales and the number of steps of noise injection. Inspired by \cite{song2020score}, a general framework based on continuous SDE based energy-based generative models is proposed for multivariate time series forecasting. 


\section{Score based generative models}
In this section, the basic theories of score matching generative models will be firstly introduced. Then two recent works that perturb data with a sequence noise scales to estimate score network are presented. The noise involvement can be seen as a discrete form of continuous-time stochastic differential equations. And then some methods from view of continuous-time SDEs for score network estimation will be introduced.
\subsection{Score matching models}
Assume $\textbf{x} \sim p_{\mathcal{X}}(\textbf{x})$ denotes the distribution of a D-dimensional dataset. Score matching \cite{hyvarinen2005estimation} is an energy-based generative model which is proposed for learning non-normalized statistical models. Instead of using maximum likelihood estimation, it aims to minimize the distance of the derivatives of the log-density function between data and model distributions. Although the density function of data distribution is unknown, the objective can be simplified as Eq. \ref{sm_obj} based on a simple trick of partial integration. 
\begin{equation}\label{sm_obj}
	\begin{aligned}
		L(\theta) &= \frac{1}{2}\mathbb{E}_{p_{\mathcal{X}}(\textbf{x})}\Vert\nabla_\textbf{x}\log p_{\theta}(\textbf{x}) - \nabla_\textbf{x}\log p_{\mathcal{X}}(\textbf{x})\Vert_2^2\\
		=& \mathbb{E}_{p_{\mathcal{X}}(\textbf{x})}[\text{tr}(\nabla_\textbf{x}^2\log p_{\theta}(\textbf{x})) + \frac{1}{2}||\nabla_\textbf{x} \log p_{\theta}(\textbf{x})||_2^2] + \text{const}
	\end{aligned}
\end{equation}
where $p_{\theta}(\textbf{x})$ represents the distribution of model estimated by neural network and $\theta$ are learnable parameters of the model. $\nabla_{\textbf{x}}\log p_{\theta}(\textbf{x})$ is called \textit{score function}. It's obvious that the optimal solution to Eq. \ref{sm_obj} equals to $\nabla_\textbf{x}\log p_{\mathcal{X}}(\textbf{x})$ for all $\textbf{x}$ and $t$.

\subsection{Discrete score matching models}
Recently, two classes of energy-based generative models that use various levels of noise to estimate score network have achieved good performance on image generation tasks. We will introduce them separately in the following part.

\subsubsection{Score matching with Langevin dynamics}
Song \cite{song2019generative} proposes score matching with Langevin dynamics (SMLD) to improve score-based generative modeling by perturbing data with various levels of noise and trains a  Noise Conditioned Score Network (NCSN) $s_{\theta}(\textbf{x}, \sigma)$ to estimate scores corresponding to all noise levels. The perturbation kernel is defined as 
\begin{equation}\label{SMLD_D}
	p_{\sigma}(\tilde{\textbf{x}}|\textbf{x}):=\mathcal{N}(\tilde{\textbf{x}}; \textbf{x}, \sigma^2\textbf{I})
\end{equation}
where $\sigma$ represents noise scales. Consider a noise sequence with ascending order $\{\sigma_1, \sigma_2, \cdots, \sigma_N\}$, where $\sigma_1$ is small enough such that $p_{\sigma_1}(\textbf{x})\approx p_{\mathcal{X}}(\textbf{x})$, $\sigma_N$ is large enough that $p_{\sigma_N}(\textbf{x})\approx \mathcal{N}(\textbf{0}, \sigma^2_N\textbf{I})$. The training process is to optimize a weighted sum of denoising score matching \cite{vincent2011connection} objective.
\begin{equation}\label{smld_loss}
	L_{\theta}= \text{argmin}_{\theta}\sum^N_{i=1}\mathbb{E}_{p_{\mathcal{X}}}\mathbb{E}_{p_{\sigma_i}}(\tilde{\textbf{x}}|\textbf{x})[\Vert s_{\theta}(\tilde{\textbf{x}}, \sigma_i)-\nabla_{\tilde{\textbf{x}}}\log p_{\sigma_i}(\tilde{\textbf{x}}|\textbf{x})\Vert_2^2]
\end{equation}

As for generation, Langevin MCMC is used for iterative sampling. Assume the number of iteration steps is M, then the sampling process for $p_{\sigma_i}(\textbf{x})$ can be formulated as
\begin{equation}\label{smld_sample}
	\textbf{x}_{i}^m = \textbf{x}_{i}^{m-1} + \epsilon_i s_{\theta}(\textbf{x}_i^{m-1}, \sigma_i) + \sqrt{2\epsilon_i}\textbf{z}_i^m, m=1,2,\cdots,M
\end{equation}
where $\textbf{z}_i^m\sim \mathcal{N}(\textbf{0}, \textbf{I})$, $\epsilon_i$ is the step size. The above iteration sampling process is repeated for N steps with $\textbf{x}_N^0\sim \mathcal{N}(\textbf{x}|\textbf{0}, \sigma_N^2\textbf{I})$ and $\textbf{x}_i^0=\textbf{x}_{i+1}^M$ when $i<N$. According to law of large numbers, $\textbf{x}_1^M$ will become an exact sample from $p_{\mathcal{X}}(x)$ if $\epsilon_i\rightarrow 0$ and $M\rightarrow \infty$

\subsubsection{Denoising diffusion probabilistic models}
\cite{sohl2015deep,ho2020denoising} construct a sequence of noise scales $0<\beta_i<1$, $i=1,2,\cdots,N$. Then a discrete Markov chain is constructed as
 \begin{equation}\label{DDPM_D}
 p(\textbf{x}_i|\textbf{x}_{i-1})\sim \mathcal{N}(\textbf{x}_i; \sqrt{1-\beta_i}\textbf{x}_{i-1}, \beta_i\textbf{I})
 \end{equation}
 
 Then the forward process can be obtained as $p(\textbf{x}_i|\textbf{x}_0)\sim\mathcal{N}(\textbf{x}_i;\sqrt{\alpha_i}\textbf{x}_0, (1-\alpha_i)\textbf{I})$, where $\alpha_i=\prod_{k=1}^i(1-\beta_k)$. As for reverse process, a variational Markov chain is constructed as $q(\textbf{x}_{i-1}|\textbf{x}_i)\sim \mathcal{N}(\textbf{x}_{i-1}; \frac{1}{\sqrt{1-\beta_i}}(\textbf{x}_i + \beta_is_{\theta}(\textbf{x}_i, i)), \beta_i\textbf{I})$. The objective can be considered as a variant of the evidence lower bound (ELBO).
\begin{equation}\label{ddpm_loss}
	\begin{aligned}
			L(\theta)&=\text{argmin}_{\theta}\sum^N_{i=1}(1-\alpha_i)  \mathbb{E}_{p_{\mathcal{X}}(\textbf{x})}\\
		&\mathbb{E}_{p_{\alpha_i}(\tilde{\textbf{x}}|\textbf{x})}[\Vert s_{\theta}(\tilde{\textbf{x}}, i)-\nabla_{\tilde{\textbf{x}}}\log p_{\alpha_i}(\tilde{\textbf{x}}|\textbf{x})\Vert^2_2]
	\end{aligned}
\end{equation}

The generation process is based on the inverse Markov chain, which depends on the estimated score network. This sampling method is called \textit{ancestral sampling} \cite{song2020score}.
\begin{equation}\label{ddpm_sample}
	\textbf{x}_{i-1}=\frac{1}{\sqrt{1-\beta_i}}(\textbf{x}_i + \beta_i s_{\theta}(\textbf{x}_i, i)) + \sqrt{\beta_i}\textbf{z}_i
\end{equation}
where $i=N, N-1,\cdots,1$

\subsection{Score matching with SDEs}
Song \cite{song2020score} points out that the noise involvement process of above two methods can be modeled as numerical form of stochastic differential equations. Without loss of generality, consider a SDE as 
\begin{equation}\label{sde}
	d\textbf{x}=f(\textbf{x},t_s)dt_s + g(t_s)d\textbf{w}
\end{equation}
where $\textbf{w}$ represents a standard Wiener process. $f(\textbf{x},t_s)$ is called \textit{drift} coefficient and $g(t_s)$ is a scalar function called \textit{diffusion} coefficient. \cite{anderson1982reverse} indicates that the reverse process of Eq. \ref{sde} also satisfies a SDE  as shown in Eq. \ref{rSDE}. Therefore, the SDE can be reversed if $\nabla_{\textbf{x}}\log p_t(\textbf{x})$ at each intermediate time step is known.
\begin{equation}\label{rSDE}
	d\textbf{x} = [f(\textbf{x},t_s)-g(t_s)^2\nabla_x \log p_{t_s}(\textbf{x})]dt_s + g(t_s)d\textbf{w}
\end{equation}

In addition, the forward process of the above two models can be treated as discrete form of continuous-time SDEs.  Specifically, Eq. \ref{SMLD_D} can be seen as discrete form a process $\{\textbf{x}(t_s)\}^1_{t_s=0}$ such that 
\begin{equation}\label{vesde}
d\textbf{x} = \sqrt{\frac{d[\sigma^2(t_s)]}{dt_s}}d\textbf{w}
\end{equation}
Note that Eq. \ref{vesde} gives a process with exploding variance when $t\rightarrow \infty$ and is called Variance Exploding (VE) SDE.

Similarly, when $N\rightarrow \infty$, Eq. \ref{DDPM_D} converges to Eq. \ref{vpsde}, which is called Variance Preserving (VP) SDE  because the variance $\boldsymbol{\Sigma}(t)$ is always bounded given $\boldsymbol{\Sigma}(0)$.
\begin{equation}\label{vpsde}
	d\textbf{x} = -\frac{1}{2}\beta(t_s)\textbf{x}dt_s + \sqrt{\beta(t_s)}d\textbf{w}
\end{equation}
Then a new type of SDE called sub-VP SDE is proposed as Eq. \ref{subvpsde}, it can be proved that the variance is always upper bounded by the corresponding VP SDE
\begin{equation}\label{subvpsde}
	d\textbf{x} = -\frac{1}{2}\beta(t_s)\textbf{x}dt_s + \sqrt{\beta(t_s)(1-e^{-2\int^{t_s}_0\beta(s)ds})}d\textbf{w}
\end{equation}

The trained score network can be used to construct reverse-time SDE and use numerical methods to sample from $p_0$. \cite{song2020score} also proposes predictor-corrector (PC) methods for sampling, detail information can be found in Appendix \ref{prd_cor}. 

\begin{figure}[t]
	\centering
	\includegraphics[height=0.13\textheight]{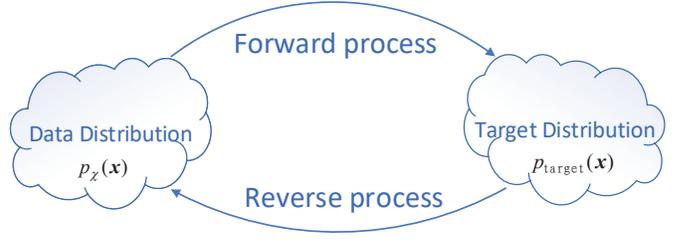} 
	\caption{Forward and reverse process of score based generative models.}
	\label{Gen}
\end{figure}

All in all, we list the relationship between these three score matching based generative models in Fig. \ref{Gen}. The forward and reverse process are listed in Table \ref{table_1}. 

\begin{table}[!t]
\caption{Comparison of DDPM, SMLD and SDE based score matching models.}
\label{table_1}
\centering
\begin{tabular}{|c|c|c|}
\hline
\diagbox{Methods}{Direction} & Forward Process & Reverse Process\\
\hline
SMLD & Eq. \ref{SMLD_D} &Eq. \ref{smld_sample} \\
\hline
DDPM & Eq. \ref{DDPM_D} & Eq. \ref{ddpm_sample}\\
\hline
SDE-Based& Eq. \ref{sde}  & Eq. \ref{rSDE} \\
\hline
\end{tabular}
\end{table}


\section{Method}
In this section, the symbols and problem definition will be firstly presented. Then, the model architecture of ScoreGrad and the network architecture of conditional score network is established. In the end, training and sampling procedure will be introduced.
\subsection{Symbol and problem formulation}
Consider a D-dimensional multivariate time series defined as $\mathcal{X}=\{\textbf{x}_1^0, \textbf{x}_2^0, \cdots, \textbf{x}_T^0\}$, where $T$ is length of $\mathcal{X}$. The probabilistic prediction tasks for multivariate time series can be converted to estimate $q_{\mathcal{X}}(\textbf{x}_{t_0:T}^0|\textbf{x}_{1:t_0-1}^0,\textbf{c}_{1:T})$, where $\textbf{c}_{1:T}$ represents covariates which are known for all time points. In this paper, an iterative forecasting strategy is adopted for given prediction time steps.
\begin{equation}\label{iter_pred}
	q_{\mathcal{X}}(\textbf{x}_{t_0:T}^0|\textbf{x}_{1:t_0-1}^0,\textbf{c}_{1:T}) = \prod_{t=t_0}^Tq_{\mathcal{X}}(\textbf{x}_t^0|\textbf{x}_{t-1}^0, \textbf{c}_{1:T})
\end{equation}

\begin{figure}[t]
	\centering
	\includegraphics[width=0.48\textwidth]{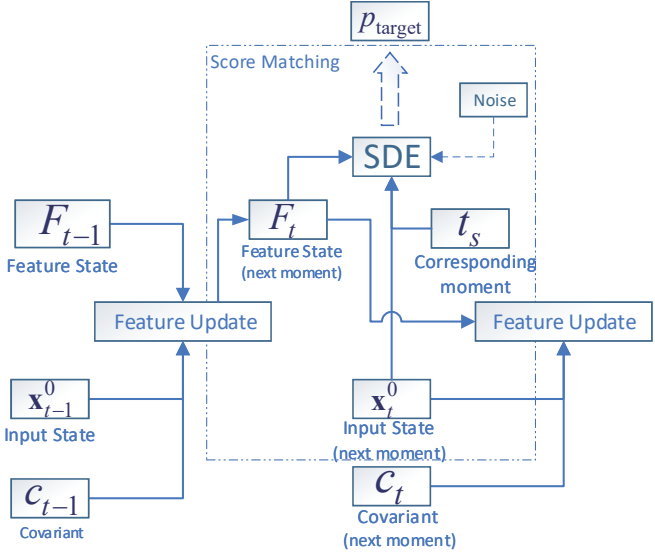} 
	\caption{Model architecture at time step t.}
	\label{Arch}
\end{figure}

\subsection{Model architecture}
The general framework of ScoreGrad can be found in Fig. \ref{Arch}. Actually, the framework can be divided into two modules at each time step:  Time series feature extraction module and conditional SDE based score matching module. Detail introduction of these two modules are presented in the following. 

\subsubsection{Time series feature extraction module} The module aims to get a feature $\textbf{F}_t$ of historical time series data until time $t-1$ and update $\textbf{F}_t$ for different time steps. The update function $R$ for $\textbf{F}_t$ can be defined as 
\begin{equation}\label{F_update}
	\textbf{F}_t = R(\textbf{F}_{t-1}, \textbf{x}_{t-1}, \textbf{c}_{t-1})
\end{equation}

 It's a general framework and many sequential modeling methods can be used here. For example, $F_t$ corresponds to hidden state $\textbf{h}_{t-1}$ if recurrent neural networks like RNN, GRU are used. For other sequential modeling methods like temporal convolutional networks (TCN) \cite{oord2016wavenet}, attention based networks \cite{vaswani2017attention}, $\textbf{F}_t$ is a vector that represents the features learned by historical data and covariates. What's more, the iterative forecasting strategy in Eq. \ref{iter_pred} can be converted to a conditional prediction problem in Eq. \ref{cond_pred}.
\begin{equation}\label{cond_pred}
	\prod_{t=t_0}^Tq_{\mathcal{X}}(\textbf{x}_t^0|\textbf{x}_{t-1}^0, \textbf{c}_{1:T}) = 	\prod_{t=t_0}^Tp_{\theta}(\textbf{x}_t^0|\textbf{F}_{t})
\end{equation}
where $\theta$ represents learnable parameters of the time series feature extraction module. In this paper, recurrent neural networks are used in ScoreGrad by default.

\subsubsection{Conditional SDE based score matching module.} 
\begin{figure}[t]
	\centering
	\includegraphics[width=0.48\textwidth]{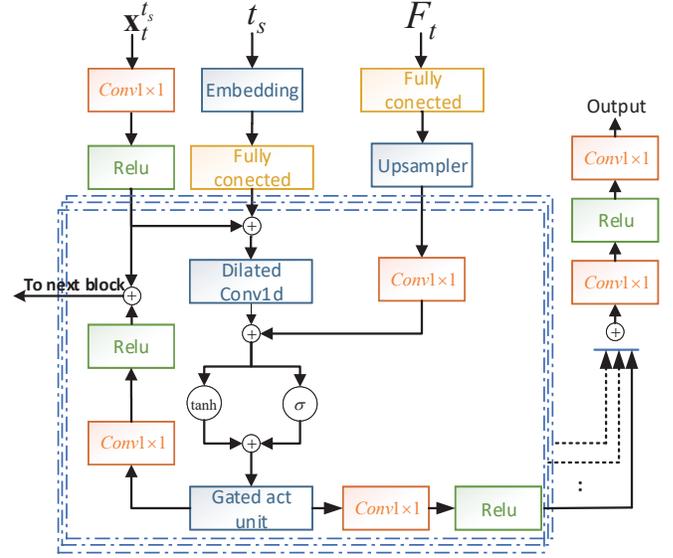} 
	\caption{Architecture of score network}
	\label{ScoreNet}
\end{figure}

As shown in Fig. \ref{ScoreNet}, $\textbf{F}_t$ is used as a conditioner for SDE based score matching models at each time step. We use the states corresponding to time t to introduce the structure of this module. The initial state distribution is $p(\textbf{x}_t^0|\textbf{F}_t)$, the forward evolve process follows Eq. \ref{sde}. Then the conditioned reverse-time SDE can be modified as
\begin{equation}\label{crsde}
	d\textbf{x}_t = [f(\textbf{x}_t,t_s)-g(t_s)^2\nabla_x \log p_{t_s}(\textbf{x}_t|\textbf{F}_{t})]dt_s + g(t_s)d\textbf{w}
\end{equation}
where $t_s\in [0, T_s]$ represents the integral time. If the conditional score function $\nabla_x \log p_{t_s}(\textbf{x}_t|\textbf{F}_{t})$ is known at each intermediate time step, the reverse continuous-time SDE can be solved and we can generate conditional samples with numerical SDE solvers. In this paper, the conditional score function is implemented with a neural network. The detail structure of the conditional score function will be illustrated in the next part.

\subsection{Conditional score network} 
Inspired by WaveNet \cite{oord2016wavenet} and DiffWave \cite{kong2020diffwave}, the conditioned score network is designed to have 8 residual blocks and the structure of a single block is shown in Fig. \ref{ScoreNet}. The input for conditioned score network includes feature $\textbf{F}_t$, input state $\textbf{x}_{t}^{t_s}$ and corresponding time $t_s$. $\textbf{x}_{t}^{t_s}$ is transformed with 1D CNNs with filter size of 1 and 3. The embedding module for $t_s$ is changed from positional embeddings \cite{vaswani2017attention} to random Fourier feature embeddings \cite{tancik2020fourier} compared with discrete form in \cite{ho2020denoising}. $\textbf{F}_t$ is served as an conditioner for score network. The sum of hidden representation of inputs and time embeddings is sent to a bidirectional dilated convolution block. Then the sum of hidden representations of conditioner and the output of dilated convolution block is sent to a gated activation unit \cite{oord2016wavenet}. Then one part of output serves as output of the block and the other one is summed up with skip-connection input and then used as the input of the next block. In the end, the output of all blocks are summed up and transformed with 1D CNN for final output.

\subsection{Training} 
In the training process, a multivariate time series data is sent to the network. At time step t, the conditioned SDE based score matching module and time series feature extraction module can be trained via a continuous form of the loss function in Eq. \ref{ddpm_loss} and Eq. \ref{smld_loss}.
\begin{equation}\label{L_t}
	\begin{aligned}
		L_t(\theta) &= \text{argmin}_{\theta}\mathbb{E}_{t_s} \big(\lambda(t_s)\mathbb{E}_{\textbf{x}_{t}^0} \mathbb{E}_{\textbf{x}_t^{t_s}|\textbf{x}_t^0} \\
		&\big[\Vert s_{\theta}(\textbf{x}_t^{t_s}, \textbf{F}_t, t_s) -\nabla_{\textbf{x}_t^{t_s}}\log p_{0t_s}(\textbf{x}_t^{t_s}|\textbf{x}_t^0)\Vert_2^2 \big]\big)\\
	\end{aligned}
\end{equation}
where $\lambda: [0, T_s]\rightarrow \mathbb{R}_{>0}$ is a weighting function, $t_s$ is randomly sampled from closed interval $[0, T_s]$. As shown by \cite{song2020score}, we can choose $\lambda \propto\mathbb{E}\big[\Vert\nabla_{\textbf{x}_t^{t_s}}\log p_{0t_s}(\textbf{x}_t^{t_s}|\textbf{x}_t^0\Vert_2^2\big]$. Then the total loss for a sequential multivariate time series $\{\textbf{x}_t^{0}, t=1,2,\cdots, T\}$ is 
\begin{equation}
	L(\theta) = \frac{1}{T}\sum^T_{t=1}L_t(\theta)
\end{equation}

All in all, the training procedure can be summarized as Algorithm \ref{train}. The loss value has a strong correlation with the type of SDE and the target distribution. In other words, the prediction performance cannot be judged according to loss value. More discussion can be found in appendix \ref{app_loss}.
\begin{algorithm}
	\caption{Training procedure of ScoreGrad}
	\label{train}
	\begin{algorithmic}
		\Require data $\textbf{x}_t^0, t=1,2,\cdots T$, integration time $T_s$
		\State Get $\textbf{F}_{1:T}$ according to Eq \ref{F_update}.
		\Repeat
		\State loss = 0
		\For{$t\leftarrow$ 1 to $T$}
		\State Initialize $t_s\sim \text{Uniform}(0, T_s)$, $\textbf{z}\sim\mathcal{N}(\textbf{0},\textbf{I})$
		\State Calculate mean $\textbf{m}_t^{t_s}$ and standard deviation $\textbf{v}_t^{t_s}$ of marginal probability $p(\textbf{x}_t^{t_s}|\textbf{x}_t^0)$ for given SDE.
		\State Get samples at $t_s$: $\textbf{x}_t^{t_s} = \textbf{m}_t^{t_s} + \textbf{z}*\textbf{v}_t^{t_s}$ 
		\State Calculate $L_t(\theta)$ according to Eq. \ref{L_t}.
		\State loss += $L_t(\theta)$
		\EndFor
		\State	loss /= T
		\State Take gradient step on $\nabla_{\theta}\text{loss}$
		\Until{converged}
	\end{algorithmic}
\end{algorithm}

\subsection{Prediction}
\begin{figure}[t]
	\centering
	\includegraphics[width=0.48\textwidth]{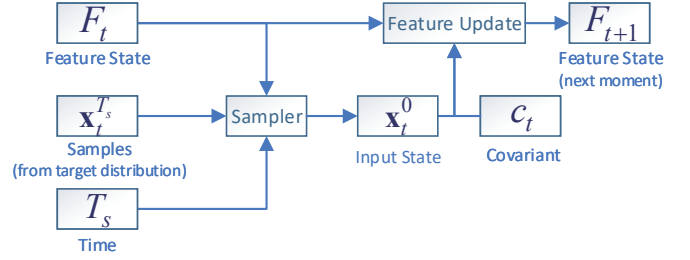} 
	\caption{Sampling procedure at time step t.}
	\label{Sampling}
\end{figure}

The prediction process can be converted into iterative sampling from the reverse continuous-time SDE. Detail sampling procedure at time step t is shown in Fig \ref{Sampling}. Firstly, get samples from target distribution and denote the samples as $\textbf{x}_t^{T_s}$. Then the feature state $\textbf{F}_t$, samples $\textbf{x}_t^{T_s}$ and corresponding time $T_s$ is sent to the sampler, which is a numerical solver of Eq \ref{rSDE}. After solving the reverse SDE, the prediction value $\textbf{x}_{t}^0$ for time step t can be obtained. Then the prediction value $\textbf{x}_{t}^0$, covariate $\textbf{c}_t$ and feature state $\textbf{F}_t$ is sent to \textit{time series feature extraction module} to get feature state at the next moment, which can be used for predicting values of next moment. Repeat the steps as above and we can get prediction values for given prediction time steps in the future.

As for sampler, it can be designed based on \textit{predictor-corrector} (PC) samplers as shown in \cite{song2020score}. The PC samplers can be regarded as a generalization of the original sampling methods. Specifically, a numerical SDE solver is used to get an initial estimation of the sample in the \textit{predictor} module. For example, Eq. \ref{smld_sample} and Eq. \ref{ddpm_sample} can be used as a predictor for continuous SMLD or DDPM models when $s_{\theta}$ is replaced by conditioned score network. Then an score-based MCMC approach is used to correct the marginal distribution of the sample in \textit{corrector} module. For example, annealed Langevin  MCMC can be used for corrector. In this framework, the sampler of SMLD can be regarded as a special PC sampler where predictor is an identity function. Similarly, the sampler of DDPM can be regarded as a PC sampler where corrector is an identity function. Detail derivation of predictor and corrector can be found in Appendix \ref{prd_cor}. All in all, the design of sampler can be summarized as Algorithm \ref{alg_sampler}. 

\begin{algorithm}
	\caption{Detail procedure of sampler at time step t.}
	\label{alg_sampler}
	\begin{algorithmic}
		\Require feature state $\textbf{F}_t$, integration time $T_s$,
		 number of time steps for reverse SDE $N$ and corrector $M$
		\State Initialize $\textbf{x}_t^{T_s}\sim p_{\text{target}}(\textbf{x})$.
		\State Divide interval $[0,T_s]$ into $N$ sections, that is $\{t_s^i\, i=1,2,\cdots N\}$.
		\For{$k\leftarrow\text{N-1}$ to 1}
		\State $\textbf{x}_t^k\leftarrow$ Predictor$(\textbf{x}_{t}^{k+1}, \textbf{F}_t, t_s^{k+1})$.
		\For{$j\leftarrow$ 1 to $M$}
		\State $\textbf{x}_t^k\leftarrow$ Corrector$(\textbf{x}_{t}^{k}, \textbf{F}_t, t_s^{k+1})$.	
		\EndFor
		\EndFor
	\end{algorithmic}
\end{algorithm}

\begin{table}[!t]
	\caption{Properties of the datasets, including dimensions, domain, frequency, time steps and prediction length.}
	\label{table_2}
	\centering
	\begin{tabular}{cccccc}
		\hline
		DATA SET&DIM.&DOM.&FREQ.&\makecell[c]{TIME\\STEPS}&\makecell{PRED.\\ LEN.}\\
		\hline
		EXCHANGE&8&$\mathbb{R}^+$&DAY&6071&30\\
		SOLAR&137&$\mathbb{R}^+$&HOUR&7009&24\\
		ELECTRICITY&370&$\mathbb{R}^+$&HOUR&5833&24\\
		TRAFFIC&963&$(0,1)$&HOUR&4001&24\\
		TAXI&1214&$\mathbb{N}$&30-MIN&1488&24\\
		WIKIPEDIA&2000&$\mathbb{N}$&DAY&792&30\\
		\hline
	\end{tabular}
\end{table}

\section{Experiments}
In this section, ScoreGrad are evaluated on six real-world datasets. Basic properties of datasets, hyperparameter settings of ScoreGrad and evaluation metrics will be firstly introduced. Several recent works for multivariate time series are used for comparison. In the end, some ablation studies are carried out for analyzing the influence of various modules on the performance of ScoreGrad.

\subsection{Datasets description and evaluation metrics}
In the experiments, \textit{Exchange} \cite{lai2018modeling}, \textit{Solar} \cite{lai2018modeling}, \textit{Electricity}, \textit{Traffic}, \textit{Taxi} and \textit{Wikipedia} are used for evaluation. Detail properties of the datasets can be found in Table \ref{table_2}. Embeddings for categorical features which describe the relationships within a category is assumed can be captured when training time series models. The preprocessing process is exactly the same as in 
\cite{salinas2019high}. In order to eliminate the impact of data value on the performance of ScoreGrad, we divide each time series by their context window mean before feeding to the model. And the samples are rescaled to the original scale in the sampling process. As for covariates $\textbf{c}_t$, time dependent embeddings (e.g. hour of day, day of week) and time independent embeddings such as lag features depending on the time frequency of the dataset are used. Furthermore, it should be noted that all covariates are known for all the periods we want to predict.

As for evaluation metrics, Continuous Ranked Probability Score (CRPS) \cite{matheson1976scoring} on each time series dimension and the sum of all time series dimensions (denoted as $\text{CRPS}_{\text{sum}}$) are employed, which are the same with those in \cite{rasul2020multi,rasul2021autoregressive}. CRPS is used to measure the compatibility of a cumulative distribution function (CDF) $F$ with an observation $x$ as
\begin{equation}\label{crps}
	\text{CRPS}(F,x)=\int_{\mathbb{R}}(F(z)-\mathbb{I}\{x\leq z\})^2dz
\end{equation}
where
$$\mathbb{I}\{x\leq z\}=\left\{
\begin{array}{rcl}
	1& x\leq z\\
	0 & \text{otherwise}
\end{array}
\right.$$

It is obvious that Eq. \ref{crps} attains minimum value when the predictive distribution $F$ is equal to the data distribution. Therefore, it's a proper scoring function for evaluation on time series forecasting tasks. Although exact CDF is intractable, it can be empirically estimated as $\hat F=\frac{1}{S}\sum^S_{s=1}\mathbb{I}\{x_s\leq z\}$, where $x_s$ represents samples from $F$. Then the CRPS value at each time step can be attained from Eq. \ref{iter_pred} according to \cite{jordan2017evaluating}. In addition, $\text{CRPS}_{\text{sum}}$ can be obtained by summing across different dimensions and then averaged over the prediction horizon, that is
\begin{equation}
	\text{CRPS}_{\text{sum}} = \mathbb{E}_t\big[\text{CRPS}(\hat F_{sum}(t), \sum_ix_{i,t}^0)\big]
\end{equation}
where integer $i\in [0, D)$ represents the $i^{th}$ dimension of time series.

Instead of using likelihood based metrics, $\text{CRPS}_{\text{sum}}$ is selected as a proper scoring function for time series forecasting tasks, which is the same with that in TimeGrad \cite{rasul2021autoregressive}.
\begin{table*}[!t]
	\caption{Comparion of $\text{CRPS}_{\text{sum}}$ of the methods on six real-world datasets (lower is better).}
	\label{table_3}
	\centering
	\scalebox{1.25}{
		\begin{tabular}{|c|cccccc|}
			\hline
			\diagbox{Method}{Dataset}&Exchange&Solar&Electricity&Traffic&Taxi&Wikipedia\\
			\hline
			VAR& \textbf{0.005}$\pm$\textbf{0.000}& 0.840$\pm$0.007& 0.038$\pm$0.003& 0.291$\pm$0.005&- &- \\
			\hline
			Lasso-VAR& 0.011$\pm$0.001&0.521$\pm$0.006 &0.026$\pm$0.000 & 0.153$\pm$0.002& -&3.142$\pm$0.007 \\
			\hline
			VES&0.006$\pm$0.000 &0.896$\pm$0.004 &0.883$\pm$0.004 &0.362$\pm$0.003 & - &- \\
			\hline
			GARCH& 0.023$\pm$0.000&0.884$\pm$0.002 &0.193$\pm$ 0.001&0.376$\pm$0.002 &- &- \\
			\hline
			KVAE&0.014$\pm$003 &0.339$\pm$0.022 &0.0502$\pm$0.019 &0.108$\pm$0.006 &- &0.095$\pm$0.012 \\
			\hline
			\makecell[c]{Vec-LSTM\\lowrank-Copula}&0.007$\pm$0.000 &0.319$\pm$0.011 &0.064$\pm$0.008 &0.103$\pm$0.006 &0.326$\pm$0.007 &0.241$\pm$0.033 \\
			\hline
			\makecell[c]{Vec-LSTM\\ind-scaling}&0.008$\pm$0.001 &0.391$\pm$0.017 &0.025$\pm$0.001 &0.087$\pm$0.041 &0.506$\pm$0.005 &0.133$\pm$0.002 \\
			\hline
			GP scaling&0.009$\pm$0.000 &0.368$\pm$0.011 &0.022$\pm$0.000 &0.079$\pm$0.000 &0.183$\pm$0.039 &1.483$\pm$1.034 \\
			\hline
			GP Copula&0.007$\pm$0.000 &0.337$\pm$0.024 &0.025$\pm$0.002 &0.078$\pm$0.002 &0.208$\pm$0.018 &0.086$\pm$0.004 \\
			\hline
			Transformer-MAF&\textbf{0.005}$\pm$\textbf{0.003}&0.301$\pm$0.014 &0.0207$\pm$0.000 &0.056$\pm$0.001 &0.179$\pm$0.002 &0.063$\pm$0.003 \\
			\hline
			TimeGrad&0.006$\pm$0.001 &0.287$\pm$0.020 &0.0206$\pm$0.001 &0.049$\pm$0.006 &0.114$\pm$0.02 &0.050$\pm$0.002 \\
			\hline
			\makecell[c]{\textbf{ScoreGrad}\\ \textbf{VP SDE}}&0.006$\pm$0.001 &\textbf{0.268}$\pm$\textbf{0.021} & \textbf{0.0192}$\pm$\textbf{0.001}&0.043$\pm$0.004 &\textbf{0.102}$\pm$\textbf{0.006} &\textbf{0.041}$\pm$\textbf{0.003} \\
			\hline
			\makecell[c]{\textbf{ScoreGrad}\\ \textbf{VE SDE}}&0.007$\pm$0.001 &0.277$\pm$0.011 &0.0199$\pm$0.001 &\textbf{0.037}$\pm$\textbf{0.003} & 0.104$\pm$0.009& 0.046$\pm$0.002\\
			\hline
			\makecell[c]{\textbf{ScoreGrad}\\ \textbf{sub-VP SDE}}&0.006$\pm$0.001 &\textbf{0.256}$\pm$\textbf{0.015} &\textbf{0.0194}$\pm$\textbf{0.001} &\textbf{0.041}$\pm$\textbf{0.004} &\textbf{0.101}$\pm$\textbf{0.004} & \textbf{0.043}$\pm$\textbf{0.002}\\
			\hline
	\end{tabular}}
\end{table*}

\subsection{Competitive methods}
Several statistical methods and deep learning based methods for multivariate time series forecasting are used as baselines for comparison study. Here we give brief introductions to these methods.
\begin{enumerate}
	\item \textbf{VAR} \cite{lutkepohl2005new} is a linear vector auto-regressive model with lag sequences based on the period of time series data. \textbf{Lasso-VAR} is similar with VAR except an extra Lasso regularization term added into the model.
	\item \textbf{VES} \cite{hyndman2008forecasting} is based on a state space model, which is also a classical auto-regressive model for sequential modeling.
	\item \textbf{GARCH} \cite{van2002go} is a multivariate conditional heteroskedastic model, which exploits unconditional information first in order to avoid convergence difficulties.
	\item \textbf{KVAE} \cite{fraccaro2017disentangled} combines Kalman filter and variational auto-encoder together and disentangles object's representation and the dynamics.
	\item \textbf{Vec-LSTM} \cite{salinas2019high} combines an RNN-based time series model with Gaussian copula process with a low-rank covariance structure, which is called \textbf{Vec-LSTM-lowrank-Copula}. In addition, we also choose another variant called \textbf{Vec-LSTM-ind-Scaling} for comparison, which outputs the parameters of an independent Gaussian distribution based on mean scaling method.
	\item \textbf{GP-Scaling} \cite{salinas2019high} unrolls an LSTM with scaling on each dimensions separately and output the distribution based on a low-rank Gaussian distribution. Different from \textbf{GP-Scaling}, \textbf{GP-Copula} obtains the joint emission distribution based on a low-rank plus diagonal covariance Gaussian copula.
	\item \textbf{Transformer-MAF} \cite{rasul2020multi} proposes to combine Transformer with Masked Autoregressive Flow \cite{papamakarios2017masked} and achieves promising results on probabilistic multivariate time-series forecasting tasks.
	\item \textbf{TimeGrad} \cite{rasul2021autoregressive} proposes to combine RNN and discrete energy-based generative models for sequential modeling and achieves state of the art results on the six datasets.
\end{enumerate}

\subsection{Hyperparameter settings}
The length of time window in the training process is set to twice the prediction length. Two layers of GRU are used for time series feature extraction module and the dimension of hidden states is set to 40 for \textit{Exchange}, \textit{Solar}, \textit{Electricity} and 128 for the other three big datasets. The integral interval of SDEs $t_s$ is set to $[0, 1]$. For VP SDE and sub-VP SDE, $\beta(t_s)$ is set as a linear transformation of $t_s$. As for VE SDE, $\sigma(t_s)$ is set as an exponential transformation of $t_s$. The range of $\beta(t_s)$ and $\sigma(t_s)$ are based on the type of SDE and dataset according to \cite{song2020improved}. $\sigma(0)$ is chosen to be as large as the maximum Euclidean distance between all pairs of training data points. In addition, we find that the sampling process is sometimes unstable in the experiments, exponential moving average (EMA) is applied to parameters to solve this issue according to Technique 5 in \cite{song2020improved}. 

As for sampling, we use predictor-corrector samplers. The predictor module can be any numerical solvers for SDE and we list three different predictors in Appendix \ref{pred_md}. The corrector module is implemented with MCMC approaches. Here we use Langevin dynamics and annealed Langevin dynamics to construct corrector. The comparison between these samplers is given in the following parts. What's more, it's worth mentioning that the number of diffusion steps in sampling is very important. In order to explore the effect of hyperparameters on the performance of model, some ablation experiments are carried out for ScoreGrad. More details about ScoreGrad can be found at \url{https://github.com/yantijin/ScoreGradPred}.

\subsection{Results}

Three different SDEs are evaluated in the framework of ScoreGrad. Table \ref{table_3} illustrates the performance of ScoreGrad and the other eleven models on the datasets. The mean and standard deviation of $\text{CRPS}_{\text{sum}}$ values are obtained by 10 runs. The top two results on each dataset are bolded in the table.

It can be known from Table \ref{table_3} that deep learning based methods generally perform better than statistical methods, especially on big datasets. Therefore, it's important to model the nonlinear dependencies of multivariate time series data. KVAE represents the methods that use autoencoder with variational inference for sequential modeling. The dimension of latent variable has a significant impact on performance. Vec-LSTM and its variants use RNN and Gaussian copula process with a low-rank covariance structure to handle non-Gaussian distribution. GP Copula performs better than KVAE on all datasets. Transformer-MAF adopts transformer to further improve the power of modeling time series and uses normalizing flows to model the distribution of time series. It performs better than GP copula on all datasets. However, the function form of Transformer-MAF is constrained. TimeGrad replaces normalizing flows with DDPM and obtains impressive results on all but the smallest dataset \textit{exchange}. However, TimeGrad is sensitive to the noise scales and number of steps for noise injection. 

DDPM is treated as a special discrete form of SDE in ScoreGrad and we propose continuous energy-based generative models for time series forecasting. It's obvious that all the three methods obtain state-of-the-art results on these datasets except the smallest dataset \textit{Exchange}. Specifically, ScoreGrad based on VP SDE performs better than ScoreGrad based on VE SDE in most cases. However, ScoreGrad based on VE SDE obtains best results on Traffic. Furthermore, the results of ScoreGrad based on sub-VP SDE are basically the same with those of ScoreGrad based on VP SDE on \textit{Electricity}, \textit{Taxi} and \textit{Wikipedia}. As for \textit{Solar}, ScoreGrad based on sub-VP SDE outperforms all the other methods. Besides, it can be known that the variance of ScoreGrad based on sub-VP SDE is slightly smaller than that of ScoreGrad based on VP SDE from table \ref{table_3}, which is consistent with the derivation.

\begin{figure*}[t]
	\centering
	\includegraphics[width=0.9\textwidth]{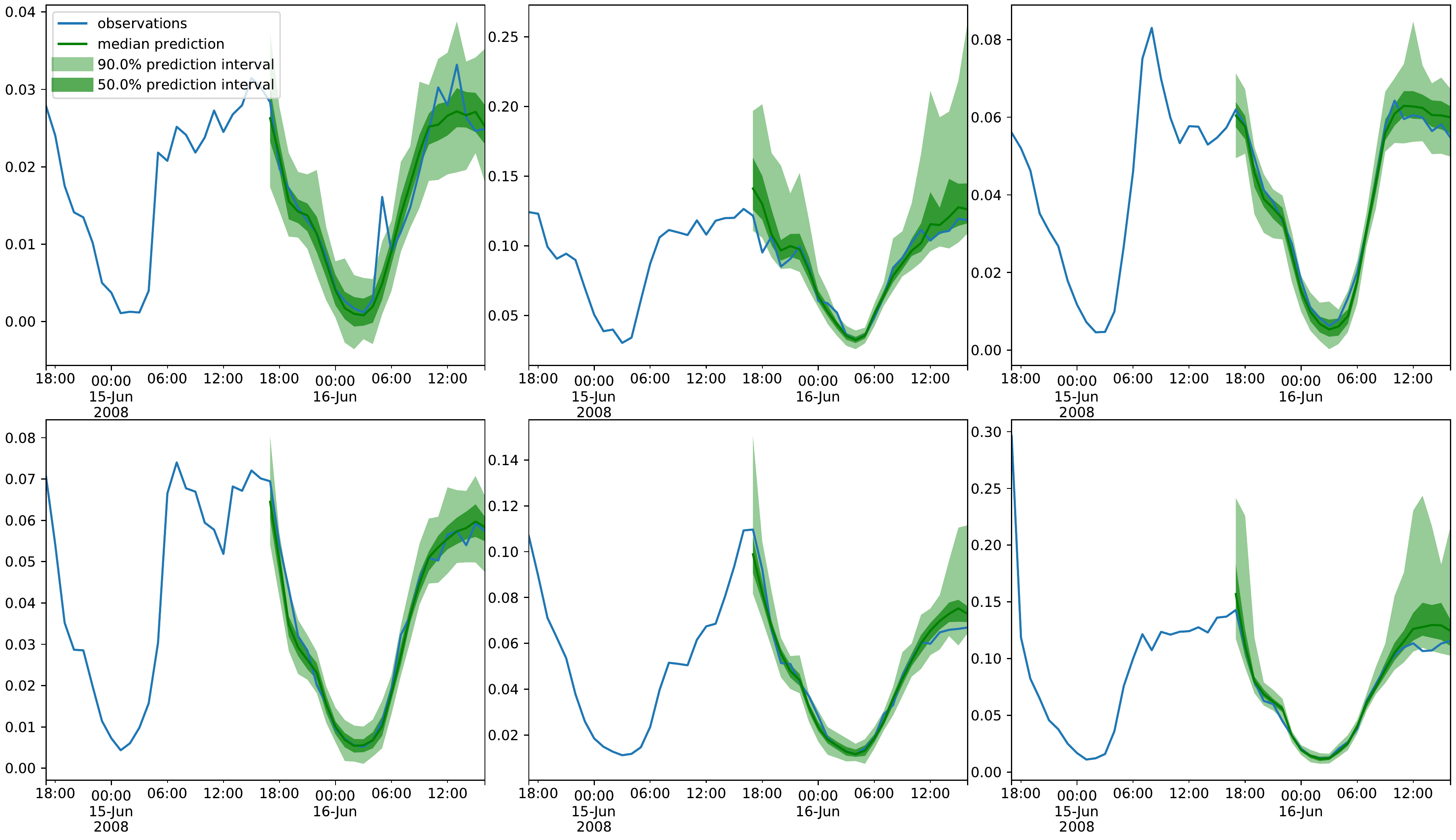} 
	\caption{Prediction intervals of ScoreGrad based on VP SDE and actual observations of Traffic.}
	\label{pred_expample}
\end{figure*}

As shown in Fig. \ref{pred_expample}, the first six channels of the test set of Traffic are used as an example of multivariate time series prediction using ScoreGrad based on VP SDE. The blue and green lines represent the observation values and median prediction. The green areas cover 50\% and 90\% distribution intervals based on VP SDE. It's obvious that most observation values are within 50\% distribution intervals. In addition, the observation values are very small and the range of values of different channels may vary greatly. Therefore, scaling all the values by dividing the context window mean is necessary for model training.

\subsection{Ablation study}
\subsubsection{Comparison of samplers}
\begin{figure}[t]
	\centering
	\includegraphics[width=0.5\textwidth]{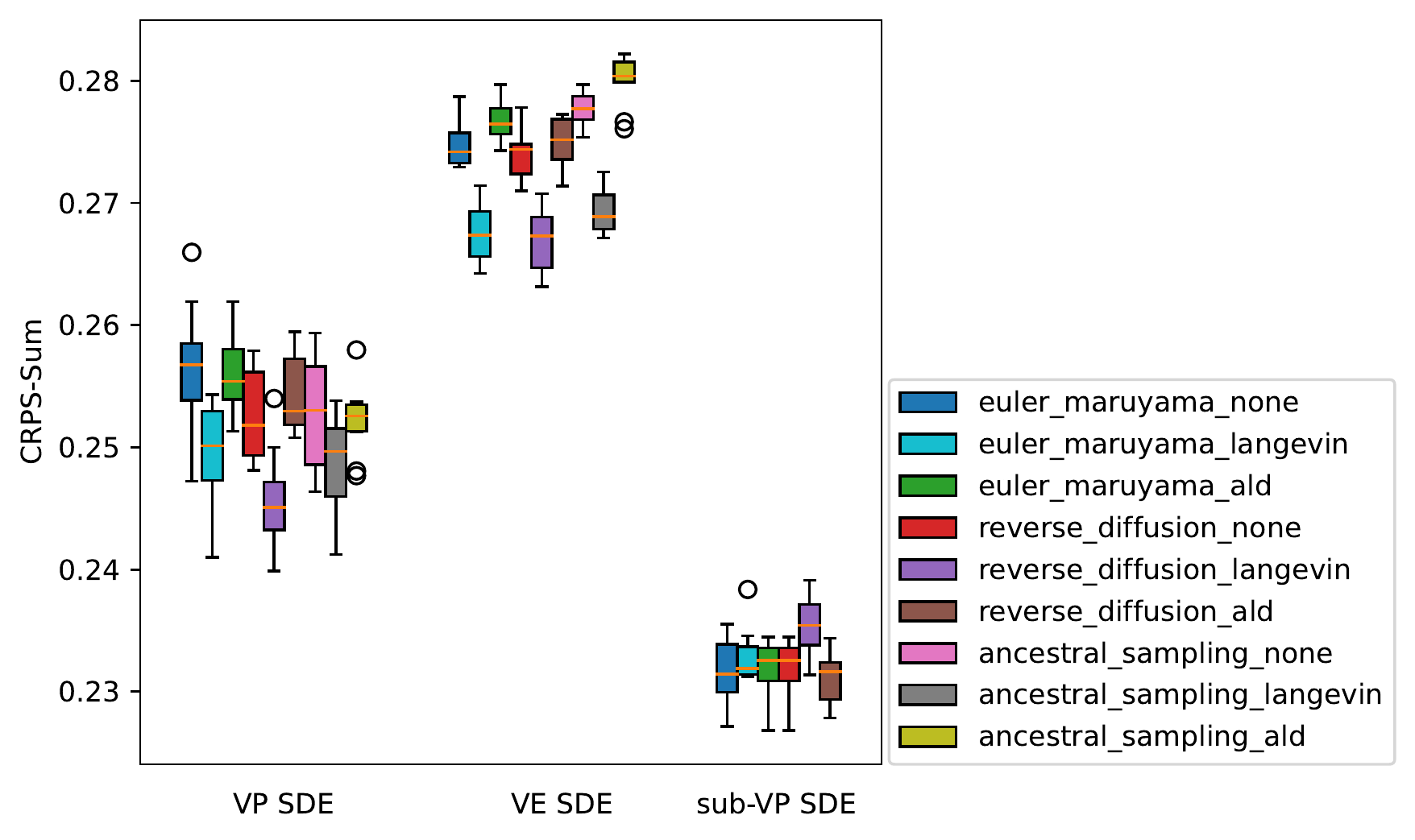} 
	\caption{Comparison of $\text{CRPS}_{\text{sum}}$ on Solar when using different samplers (lower is better).}
	\label{comp_sampler}
\end{figure}

In order to explore the impact of the type of samplers on the performance of prediction, we firstly load a trained model and then change the type of samplers while keeping the other parameters unchanged. In this paper, there are three different predictors and two different correctors. Therefore, there are six different samplers. Besides, we also compare the performance of PC samplers with or without corrector. All in all, there are nine different samplers for a certain trained model. The performance of these samplers based on three trained models on \textit{Solar} is shown in Fig. \ref{comp_sampler}. 

It's worth mentioning that we can not directly compare the performance between three models in Fig. \ref{comp_sampler} because they only represent the result of a certain training, not the average of multiple training.  For ScoreGrad based on VP SDE, PC samplers with corrector performs better than those without corrector. However, PC samplers using annealed Langevin dynamics as corrector performs slightly worse than those without corrector for ScoreGrad based on VE SDE for \textit{Solar}. Furthermore, the combination of reverse diffusion sampler and Langevin dynamics performs best on VP SDE and VE SDE. In addition, it can be known that reverse diffusion sampler always outperforms ancestral sampling on this dataset. Besides, the variance of sub-VP SDE is smaller than that of VP SDE, which is consistent to the derivation. Ancestral sampling is not feasible for sub-VP SDE. As for the remaining six samplers, the combination of reverse diffusion sampler and annealed Langevin dynamics performs best. The comparison of samplers on the other five datasets can be found in Appendix \ref{cp_s}.

\subsubsection{Effect of diffusion steps}
\begin{figure}[t]
	\centering
	\includegraphics[width=0.49\textwidth]{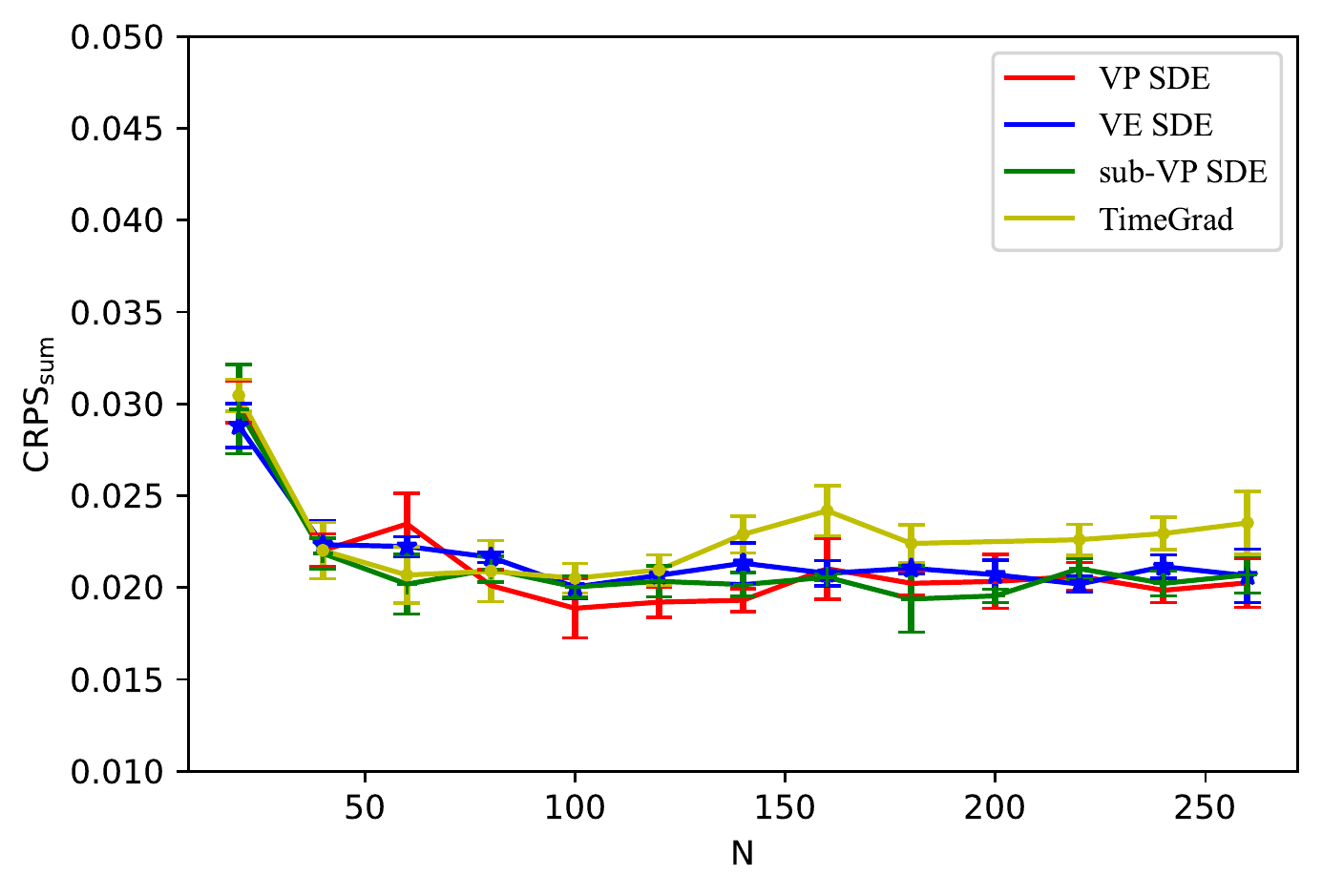} 
	\caption{Comparison of $\text{CRPS}_{\text{sum}}$ of ScoreGrad and TimeGrad on the test set of \textit{Electricity} data by varying the number of steps of noise injection (lower is better).} 
	\label{scale_comp}
\end{figure}

The number of diffusion steps $N$ in ScoreGrad corresponds to the number of noise injection steps in DDPM. TimeGrad is sensitive to the number of steps of noise injection. In contrast, the forward process of ScoreGrad doesn't depend on the diffusion steps because it employs a continuous energy-based generative model and can get the state at any moment. Therefore, it avoids the problem in TimeGrad. Actually, the number of diffusion steps is only used in the sampling process in ScoreGrad. In order to explore the effect of the diffusion steps on the performance of the model, we compare $\text{CRPS}_{\text{sum}}$ of \textit{Electricity} by varying the diffusion or noise injection steps of ScoreGrad and TimeGrad. Specifically, we set $N=20,40,\cdots, 260$ and obtain the mean and standard error metrics by retraining and evaluating 5 times for each $N$. The results are shown in Fig. \ref{scale_comp}.

The performance of ScoreGrad is basically the same with that of TimeGrad when $N=20$. In addition, it's worth mentioning that the performance of ScoreGrad and TimeGrad are already better than that of KVAE and Vec-LSTM with low rank Copula process when $N=20$, which indicates that energy-based models are suitable for modeling time series. With the increase of $N$ when $N\leq 100$, the performance of TimeGrad becomes better. The optimal value of noise injection steps $N$ is approximately 100 for TimeGrad. Although larger $N$ allows the reverse process to be approximately an Gaussian for TimeGrad according the derivation above, larger $N$ will hurt the performance. ScoreGrad based on VP SDE and VE SDE obtain best result when $N=100$ while $N=180$ for sub-VP SDE. Furthermore, all the three methods of ScoreGrad perform better than TimeGrad when $N\geq 100$. $\text{CRPS}_{\text{sum}}$ values of these three methods fluctuate in a small range with the increase of $N$. We also find similar behavior with the other datasets. Therefore, the performance of ScoreGrad based on VP SDE, VE SDE and sub-VP SDE are more robust to the changes of $N$ compare to TimeGrad. Continuous energy-based generative models can avoid setting the number of injection steps in the training process, which is important in numerical form of SDEs as shown in TimeGrad.


\subsection{Other applications}
\begin{figure}[t]
	\centering
	\includegraphics[width=0.49\textwidth]{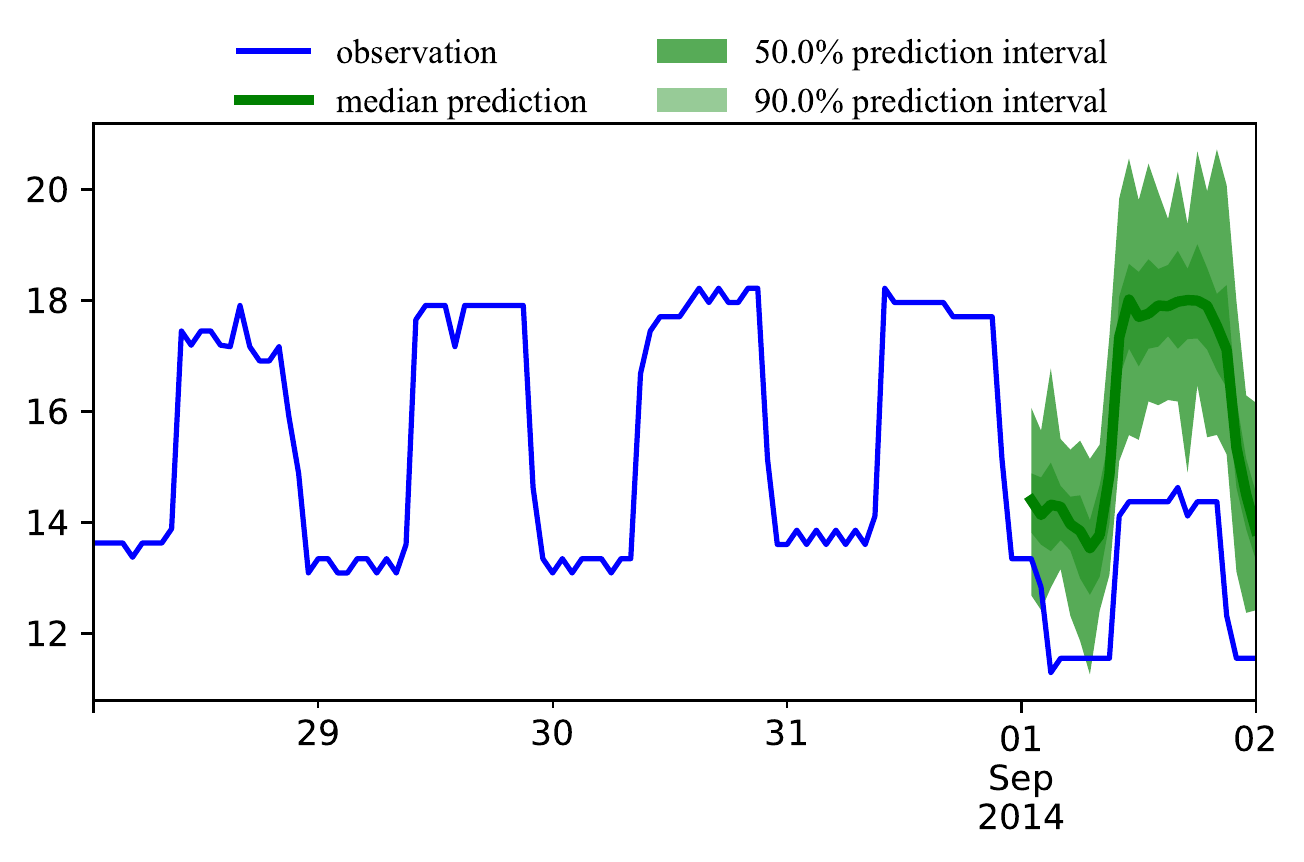} 
	\caption{The prediction of a single channel of \textit{Electricity} based on ScoreGrad. } 
	\label{abnormal}
\end{figure}

When we plot prediction values based on ScoreGrad, we find the values of some channels do not follow history patterns. For example, Fig. \ref{abnormal} illustrates the prediction intervals of a simple channel of \textit{Electricity} based on ScoreGrad. It's obvious that the data exhibits the periodic property. However, the values decrease suddenly on September 1st. There are many probabilities for this phenomenon, such as the difference of patterns between holidays and weekdays, power failure in local area. In any case, there should be real-time monitoring to ensuring the reliability of service. 

ScoreGrad can be also used for time series anomaly detection. For example, if observation values are not in the area of 90\% prediction intervals for some time (e.g., one hour), the alarm information should be sent to concerned people for more attention. It's obvious that the values on September 1st will get more attention under such a rule.

\section{Conclusion and future work}
In this paper, we propose ScoreGrad, a multivariate probabilistic time series forecasting framework that adopts continuous energy-based generative models to learn the distribution of data. The framework can be divided into time series feature extraction module and conditional SDE based score matching module. Compared with VAE or flow based models, EBMs have few constrains on function forms. In addition, the connection between EBMs and SDEs enable EBMs to be built within a broader framework. ScoreGrad achieves state-of-the-art results on six real world datasets. What's more, the ablation studies and analysis of samplers indicate that ScoreGrad can better model multivariate time series compared with existing works.

There is still much work to be done based on the proposed framework. The first direction is about improvement of the network architecture. Powerful sequential modeling methods like Transformer can be used for time series feature extraction module. More SDEs can be designed for conditional SDE based score matching module. The second direction is about reducing computation cost of sampling. \cite{song2020denoising} proposes to use non-Markov process which allowed for fast sampling. \cite{shen2019randomized} proposes a fast sampling method for log-concave sampling based on underdamped Langevin diffusion. The third direction is to establish the connection between loss and metrics. The loss function  used in this paper is not directly related to the prediction performance, which is not friendly for the design and evaluation of new models based on the framework. 
We leave these three directions as future work.


%

\appendices
\section{Train loss of ScoreGrad}\label{app_loss}
\begin{figure}[t]
	\centering
	\includegraphics[width=0.49\textwidth]{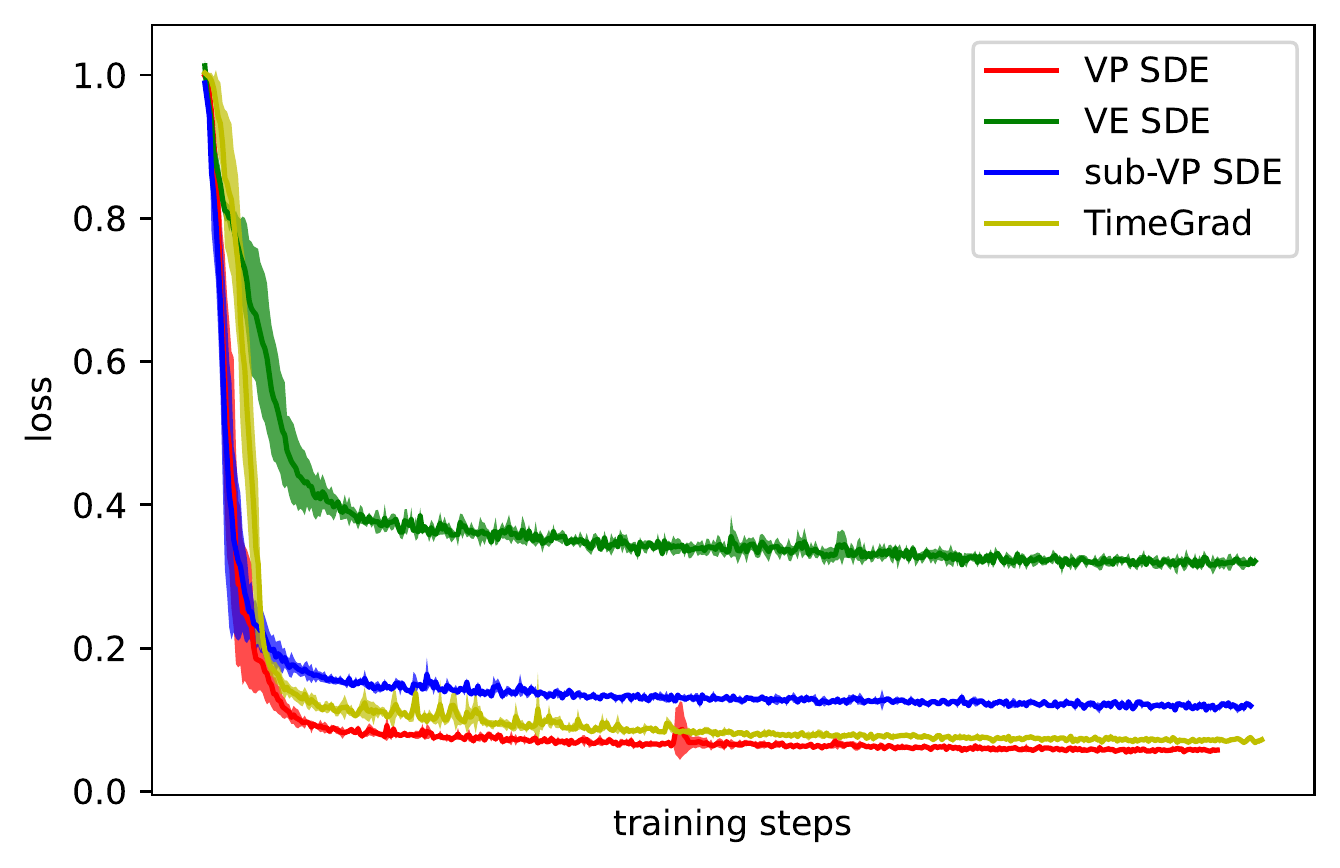} 
	\caption{Train loss of ScoreGrad and TimeGrad on Electricity.} 
	\label{loss}
\end{figure}

The prediction performance can not be judged by the loss value. For example, Fig. \ref{loss} illustrates the training loss of ScoreGrad and TimeGrad on Electricity. Each curve is obtained by training 5 times. It can be known that although the loss value of TimeGrad is smaller than that of ScoreGrad based on sub-VP SDE and VE SDE, the performance in table \ref{table_3} is not consistent with loss value. Therefore, it's not convenient to design new SDEs for ScoreGrad. We leave it as future work to establish the connection between metrics and training loss as future work.

\section{Predictor-Corrector Sampler}\label{prd_cor}
As noted in \cite{song2020score}, the predictor can be any numerical solver for the reverse-time SDE with a fixed discretization strategy. The corrector can be implemented with any score-based MCMC approach. 

\subsection{Predictor methods}\label{pred_md}
General numerical solver for SDEs such as Euler-Maruyama and stochastic Runge-Kutta methods can be used as predictor. Without loss of generality, given an reverse-time SDE as Eq. \ref{crsde}, the Euler-Maruyama methods can be formulated as
\begin{equation}\label{euler}
	\begin{aligned}
		\textbf{x}_{t}^{i} &= \textbf{x}_{t}^{i+1} - g(t_{i+1})\sqrt{\delta} \textbf{z}_{i+1}\\
		 &- (f(\textbf{x}_t^{i+1}, t_{i+1}) - g(t_{i+1})^2 s_{\theta}^*(\textbf{x}_t^{i+1}, t_{i+1}, \textbf{F}_t)) \delta
	\end{aligned}
\end{equation}
where $\delta$ represents step size, $i=1,2,\cdots, N$, $\textbf{z}_{i+1}$ follows standard Gaussian distribution.	 In addition, ancestral sampling method in Eq. \ref{ddpm_sample} has been proved to be a special discretization to the same reverse-time SDE \cite{song2020score}.
\begin{equation}\label{ances_vp}
	\begin{aligned}
	\textbf{x}_t^{i-1}&=\frac{1}{\sqrt{1-\beta_i}}(\textbf{x}_t^i + \beta_i s_{\theta}(\textbf{x}_t^i, i, \textbf{F}_t)) + \sqrt{\beta_i}\textbf{z}_i\\
	&=(2-\sqrt{1-\beta_{i}})\textbf{x}_t^{i} + \beta_{i}s_{\theta}(\textbf{x}_t^i, i, \textbf{F}_t) + \sqrt{\beta_{i}}\textbf{z}_i
	\end{aligned}
\end{equation}

The ancestral sampling method for SMLD can be similarly derived as 
\begin{equation}\label{ances_ve}
	\textbf{x}_t^{i-1} = \textbf{x}_t^{i} + (\sigma_i^2 - \sigma_{i-1}^2) s_{\theta}^*(\textbf{x}_t^{i}, i, \textbf{F}_t) + \sqrt{\frac{\sigma_{i-1}^2(\sigma_i^2-\sigma_{i-1}^2)}{\sigma_i^2}} \textbf{z}_{i}
\end{equation}

Deriving the ancestral samplers for new SDEs can be non-trivial. For example, ancestral sampling is not feasible for sub-VP SDE. Song \cite{song2020score} also proposes a new sampler called \textit{reverse diffusion sampler}, which can be used here with minor modification
\begin{equation}\label{reverse_diff}
	\begin{aligned}
		\textbf{x}_{t}^{i} =&~~ \textbf{x}_{t}^{i+1}  + g(t_{i+1})^2 s_{\theta}^*(\textbf{x}_t^{i+1}, t_{i+1}, \textbf{F}_t)\\
		& + g(t_{i+1})\textbf{z}_{i+1} - f(\textbf{x}_t^{i+1}, t_{i+1})
	\end{aligned}
\end{equation}
where $\textbf{z}_{i+1}$ follows standard Gaussian distribution.

All in all, the predictor in Algorithm \ref{alg_sampler} can be set as Eq. \ref{euler}, Eq. \ref{reverse_diff} or ancestral sampling methods in Eq. \ref{ances_ve} and Eq. \ref{ances_vp} based on the type of SDEs.

\subsection{Corrector methods}
MCMC methods such as Hamiltonian Monte Carlo can be used as corrector. In this paper, we use Langevin dynamics for sampling. In addition, annealed Langevin dynamics with minor modifications can also be used for corrector module and detail procedure for VE SDE and VP SDE can be found in Algorithm \ref{corr_VE} and Algorithm \ref{corr_VP}. 

As for sub-VP SDE, there are no similar discrete form as VP SDE. Therefore, $\alpha_i$ is set as 1 for computing step size when Algorithm \ref{corr_VP} is used for sub-VP SDE. $r$ is called "signal-to-noise" ratio and is set as 0.16 in the experiments.

\begin{algorithm}
	\caption{Corrector algorithm at each time step for VE SDE.}
	\label{corr_VE}
	\begin{algorithmic}
		\Require $\{\sigma_i\}_{i=1}^N, r,N, M$
		\State $\textbf{x}_N^{0}\sim \mathcal{N}(\textbf{0}, \sigma_{\text{max}}^2\textbf{I})$
		\For{$i\leftarrow\text{N}$ to 1}
		\For{$j\leftarrow$ 1 to $M$}
		\State $\textbf{z}\sim \mathcal{N}(\textbf{0}, \textbf{I})$
		\State $\textbf{g}\leftarrow \textbf{s}_{\theta}(\textbf{x}_i^{j-1}, \sigma_i, \textbf{F}_t)$
		\State $\epsilon\leftarrow 2(r\Vert\textbf{z}\Vert_2/\Vert \textbf{g}\Vert_2)^2$
		\State $\textbf{x}_{i}^j\leftarrow \textbf{x}_i^{j-1} + \epsilon \textbf{g} + \sqrt{2\epsilon}\textbf{z}$
		\EndFor
		\State $\textbf{x}_{i-1}^0\leftarrow\textbf{x}_i^M$
		\EndFor\\
		\Return $\textbf{x}_0^0$
	\end{algorithmic}
\end{algorithm}

\begin{algorithm}
	\caption{Corrector algorithm at each time step for VP SDE.}
	\label{corr_VP}
	\begin{algorithmic}
		\Require $\{\beta_i\}_{i=1}^N, \{\alpha_i\}_{i=1}^N r,N, M$
		\State $\textbf{x}_N^{0}\sim \mathcal{N}(\textbf{0}, \textbf{I})$
		\For{$i\leftarrow\text{N}$ to 1}
		\For{$j\leftarrow$ 1 to $M$}
		\State $\textbf{z}\sim \mathcal{N}(\textbf{0}, \textbf{I})$
		\State $\textbf{g}\leftarrow \textbf{s}_{\theta}(\textbf{x}_i^{j-1}, i, \textbf{F}_t)$
		\State $\epsilon\leftarrow 2\alpha_i(r\Vert\textbf{z}\Vert_2/\Vert \textbf{g}\Vert_2)^2$
		\State $\textbf{x}_{i}^j\leftarrow \textbf{x}_i^{j-1} + \epsilon \textbf{g} + \sqrt{2\epsilon}\textbf{z}$
		\EndFor
		\State $\textbf{x}_{i-1}^0\leftarrow\textbf{x}_i^M$
		\EndFor\\
		\Return $\textbf{x}_0^0$
	\end{algorithmic}
\end{algorithm}
\section{Comparison of samplers}\label{cp_s}
\begin{figure*}[t]
	\centering
	\includegraphics[width=0.88\textwidth]{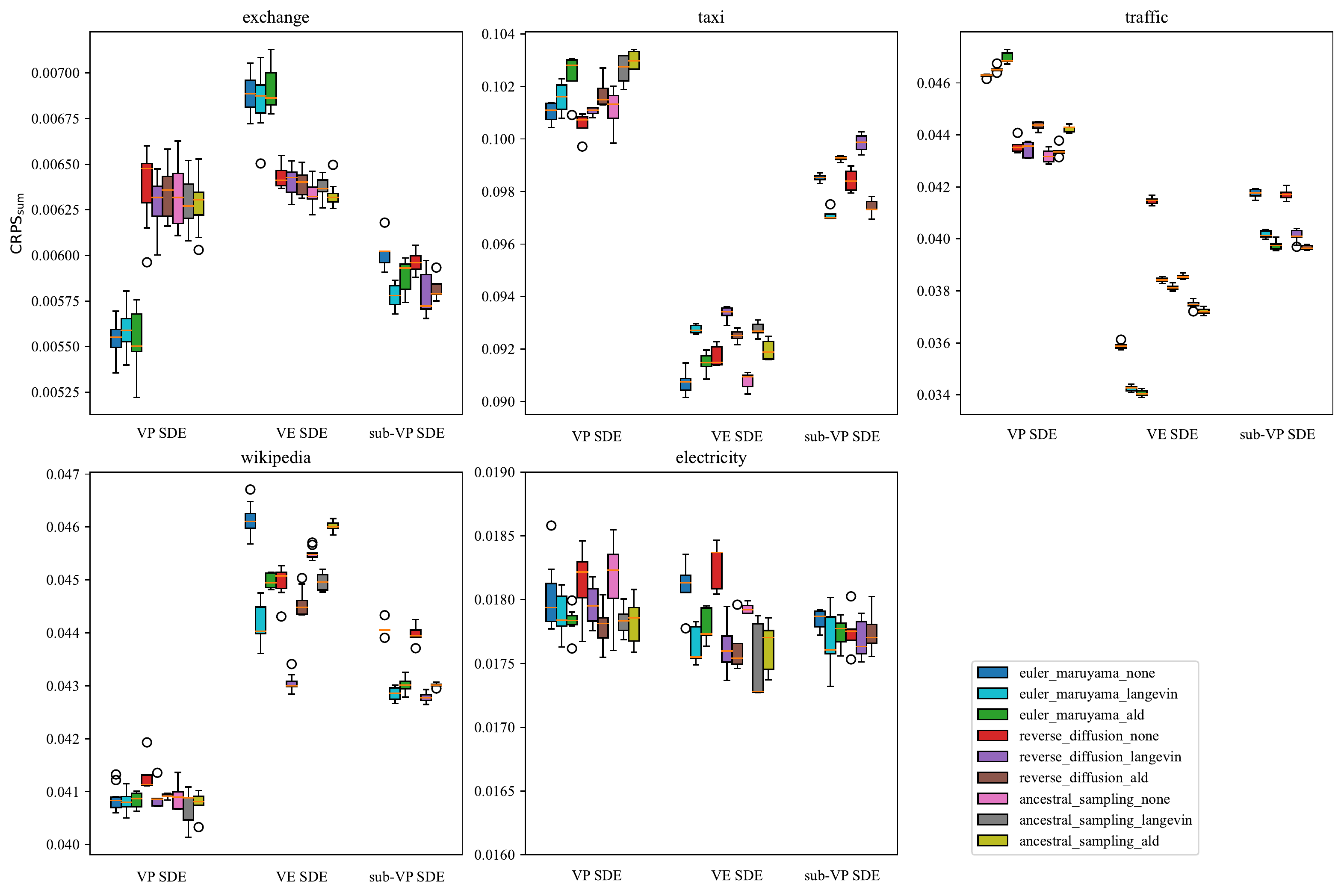} 
	\caption{Comparison of samplers on the other five datasets. } 
	\label{5_plot}
\end{figure*}

The performance of the model on the other five datasets when using different samplers for evaluation are shown in Fig. \ref{5_plot}.

%
%
%


\ifCLASSOPTIONcaptionsoff
  \newpage
\fi



%




\bibliographystyle{IEEEtran}
\bibliography{ref}
\end{document}